\documentclass[10pt,twocolumn,letterpaper]{article}

\usepackage{iccv}

\usepackage{times}
\usepackage{epsfig}
\usepackage{graphicx}
\usepackage{grffile}
\usepackage{amsmath}
\usepackage{amssymb}
\usepackage{booktabs}
\usepackage{mwe}
\usepackage{colortbl}
\usepackage{multirow}
\usepackage[ruled,vlined]{algorithm2e}

\usepackage{enumitem}
\setlist{nosep}

\usepackage{color}

\definecolor{mygray}{rgb}{0.5, 0.5, 0.5}

\newcommand{\img}{I}
\newcommand{\flow}{F}

\newcommand{\Bg}{\text{Bg}}
\newcommand{\syntFg}{\overline{\text{Fg}}}

\newcommand{\syntflow}{\overline{\flow}}
\newcommand{\mask}{M}

\newcommand{\comb}{{\bf C}}

\newcommand{\calL}{\mathcal{L}}

\SetKwInput{KwInput}{Input}
\SetKwInput{KwOutput}{Output}

\newcommand{\given}{|}

\DeclareMathOperator*{\argmin}{arg\,min}

\usepackage[breaklinks=true,bookmarks=false,colorlinks=true,linkcolor=red,citecolor=blue]{hyperref}

\iccvfinalcopy 

\def\iccvPaperID{****} 

\ificcvfinal\fi

\begin{document}

\title{Learning to Better Segment Objects from Unseen Classes \\ with Unlabeled Videos}

\author{Yuming Du \qquad \quad Yang Xiao \qquad \quad Vincent Lepetit
\\
LIGM, Ecole des Ponts, Univ Gustave Eiffel, CNRS, Marne-la-vallée, France\\
{\tt\small \{yuming.du, yang.xiao, vincent.lepetit\}@enpc.fr}\\
\url{https://dulucas.github.io/Homepage/gbopt/}}

\maketitle
\ificcvfinal\thispagestyle{empty}\fi

\begin{abstract}
 The ability to localize and segment objects from unseen classes would open the door to new applications, such as autonomous object learning in active vision.  Nonetheless, improving the performance on unseen classes requires additional training data, while manually annotating the objects of the unseen classes can be labor-extensive and expensive. In this paper, we explore the use of unlabeled video sequences to automatically generate training data for objects of unseen classes. It is in principle possible to apply existing video segmentation methods to unlabeled videos and automatically obtain object masks, which can then be used as a training set even for classes with no manual labels available.  However, our experiments show that these methods do not perform well enough for this purpose.  We therefore introduce a Bayesian method that is specifically designed to automatically create such a training set: Our method starts from a set of object proposals and relies on (non-realistic) analysis-by-synthesis to select the correct ones by performing an efficient optimization over all the frames \textbf{simultaneously}.  Through extensive experiments, we show that our method can generate a high-quality training set which significantly boosts the performance of segmenting objects of unseen classes. We thus believe that our method could open the door for open-world instance segmentation using abundant Internet videos. 
\end{abstract}
\vspace{-5mm}

\section{Introduction}
\label{sec:introduction}

\begin{figure}
    \setlength{\tabcolsep}{1.5pt}
    \includegraphics[width=.5\textwidth]{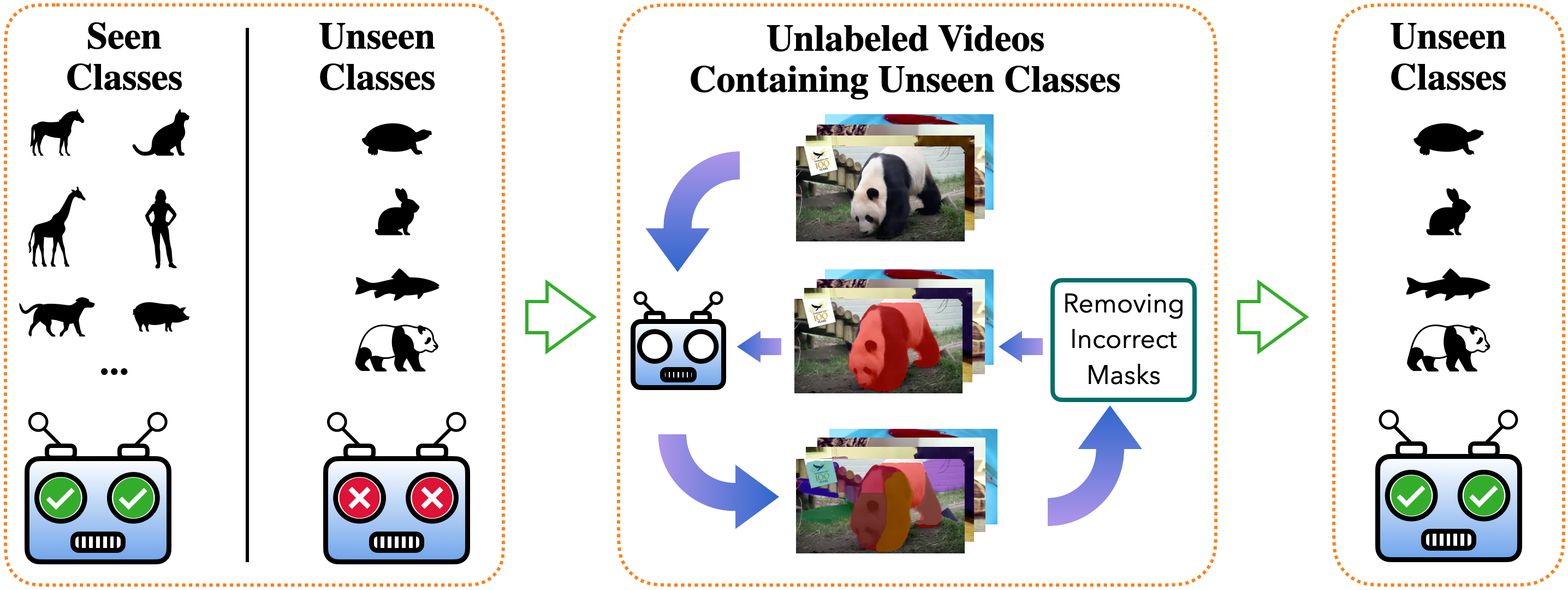}
    \caption{ Starting from an instance segmentation model trained on some classes, we want to learn to localize and segment objects from new classes without any human label. We do this by using unlabeled videos, which are an abundant source of data. Our approach can automatically detect and select the object masks in the videos. We then use the selected masks to retrain the initial model that can then localize and segment objects from the new classes in still frames without losing performance on the old ones. }
    \label{fig:teaser_robot}
 \end{figure}

Instance segmentation models are now able to predict the masks of objects of known classes in query images~\cite{he2017mask,tian2020conditional,wang2020solo}, providing rich information for many downstream applications such as scene understanding~\cite{gkioxari2019mesh,Runz_2020_CVPR} and robot grasping~\cite{wada2019joint, xie2020unseen, yen2020learning}.
Unfortunately, existing instance segmentation methods perform poorly on new classes~\cite{dhamija2020overlooked}. This is an obstacle to the development of autonomous systems evolving in open worlds where there will always be objects that do not belong to known classes.  Being able to detect and segment these objects would be the starting point of learning to grasp and manipulate them, for example.

As Figure~\ref{fig:teaser_robot} illustrates, our goal is therefore to automatically improve the performance of instance segmentation models on static images containing objects from new classes without human intervention. This is in contrast with previous works that aiming at limiting the manual labeling burden for object segmentation by using bounding boxes only~\cite{hu2018learning, Kuo2019ShapeMaskLT, Zhou2020LearningSP} or developing few-shot techniques~\cite{Fan2020FGNFG,Yan2019MetaRT}, but still require human intervention for new classes.

More specifically, we do \emph{not} aim at predicting the categories of these new objects, but focus on robustly localizing and accurately segmenting them. In this sense, our work is therefore more related to recent object discovery methods, which attempt to segment objects without manual segmentation labels by grouping pixels according to some criterion~\cite{burgess2019monet,greff2019multi,slot_attention,pham2018bayesian,wong2020identifying}.  However, these approaches are still very fragile as they can be easily affected by low-level perturbations in color, lighting, or texture.



\begin{figure}
    \setlength{\tabcolsep}{1.5pt}
    {\renewcommand{\arraystretch}{.7}
    \begin{tabular}{c c}
     \includegraphics[trim=0 1.5cm 0 0,clip,width=.24\textwidth]{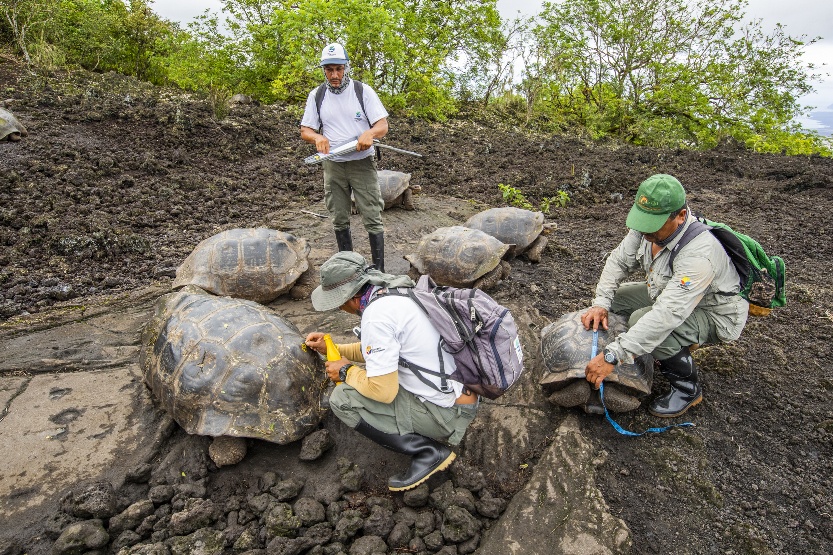}\hfill 
     & \includegraphics[trim=0 1.5cm 0 0,clip,width=.24\textwidth]{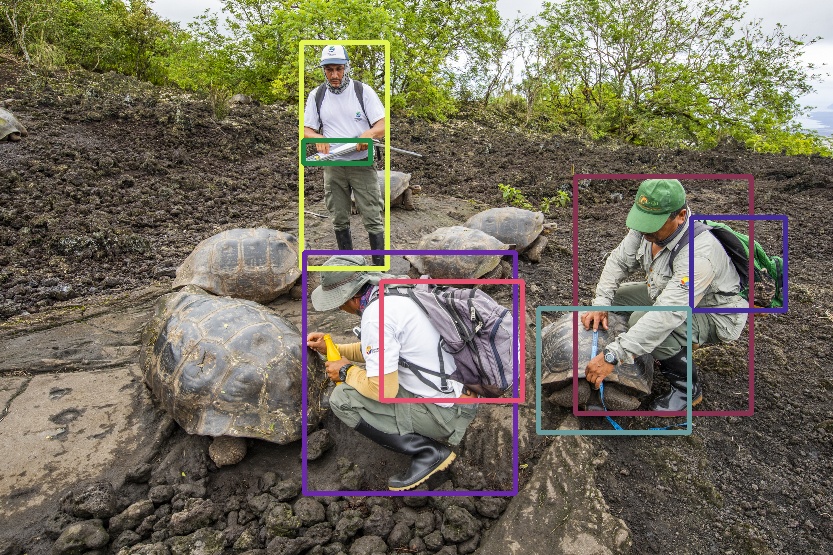}\hfill
     \\
     Input image & Confidence score $> 0.5$ \\
     \includegraphics[trim=0 1.5cm 0 0,clip,width=.24\textwidth]{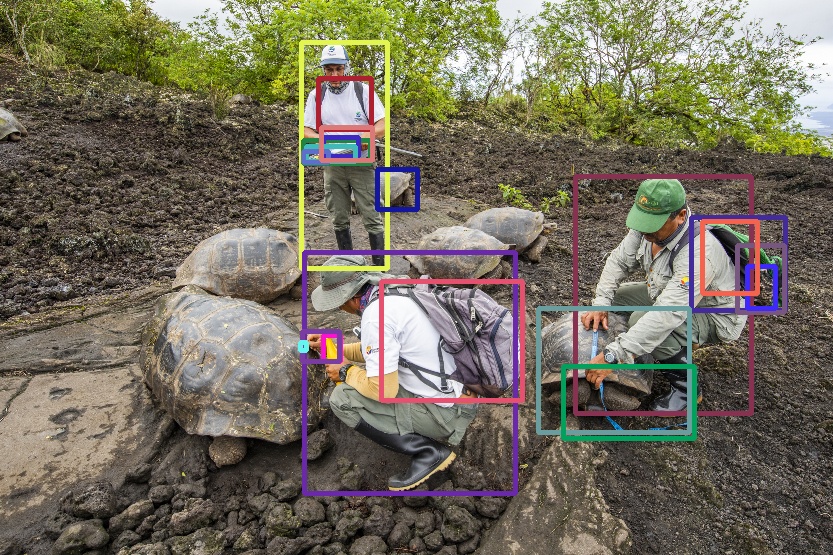}\hfill 
     & \includegraphics[trim=0 1.5cm 0 0,clip,width=.24\textwidth]{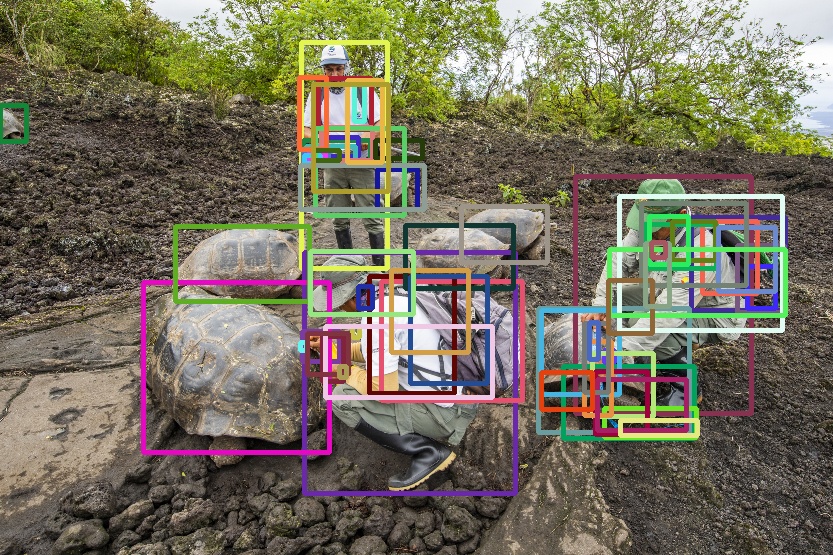}\hfill
     \\
     Confidence score $> 0.1$ & Confidence score $> 0.01$ \\
    \end{tabular}
    }
    \caption{The class 'tortoise' does not belong to COCO classes. By decreasing the confidence score threshold of Mask R-CNN~\cite{he2017mask} trained on COCO, we can eventually localize and segment all the tortoises at the price of introducing many false positives.  We filter these false positives using unlabeled video data. }
    \label{fig:teaser}
    \vspace{-4mm}
 \end{figure}
 
Our strategy is therefore to rely on unlabeled video sequences, as unlabeled videos can be acquired without effort while providing rich information.  The general idea of using videos for self-supervision is not new~\cite{dave2019towards, jakab20self-supervised, uvc_2019, lu2020learning, wang2019learning, watters2017visual}. Here, we consider video sequences to automatically generate masks for the objects visible in their frames. Using these masks for training an instance segmentation model should then make it perform better on new objects visible in the videos, even if no human intervention was provided in the process. In our experiments, we still provide the videos, but it is possible to imagine a system that captures videos by itself.

Unfortunately, our early experiments showed that state-of-the-art video segmentation methods~\cite{faktor2014video, luiten2020unovost, papazoglou2013fast, uvc_2019, ventura2019rvos} were not sufficient for this purpose.  We therefore developed our own method for automatically creating object masks from videos.  Note that our task is subtly different from video segmentation methods, which aim at keeping track of the objects over consecutive frames.  For our own goal, this is not needed and we only focus on correctly localizing and segmenting the objects in each frame.  Like some video segmentation methods~\cite{luiten2020unovost}, we start from mask hypotheses for the objects visible in the videos obtained with a pre-trained class-agnostic instance segmentation. Such model generalizes to objects from unseen classes to some extent, but at the price of introducing many false positives~\cite{ovsep20204d}. Figure~\ref{fig:teaser} shows that when the confidence threshold is lowered, both the detections of objects from unseen classes and incorrect detections get accepted.

It is however possible to filter the incorrect masks using information provided by the videos. Some methods~\cite{cheng2018fast, li2013video, luiten2020unovost, milan2015joint, Perazzi_2015_ICCV} rely on tracklets to track and filter the masks.  We claim that such strategy is not optimal, especially for our goal: Image background is underexploited, while it can be very useful by indicating the \emph{absence} of objects; The optical flow is not used or underexploited, while it gives strong cues about moving objects in videos. To fully exploit the background and motion information across unlabeled video frames, we therefore developed a (non-realistic) analysis-by-synthesis approach. 
Using a Bayesian framework, we derive an objective function with an additional non-overlapping constraint. The objective function, which consists of three loss terms, is designed to explore the background and motion information to remove incorrect masks and select masks that are temporally consistent over the entire video. The non-overlapping constraint comes from the fact that one pixel can at most belong to one object in the image and helps to reject some false positives. Moreover, we provide a two-stage optimization algorithm to optimize this objective function efficiently.


To evaluate our approach, we created a novel dataset named Unseen-VIS based on the YouTube Video Instance Segmentation~(YouTube-VIS) dataset~\cite{yang2019video}, which contains objects that do \emph{not} belong to COCO classes. Starting from the raw masks generated on the training part of Unseen-VIS using a class-agnostic Mask R-CNN pretrained on COCO dataset, we apply our method to automatically select the correct masks. We demonstrate that using these masks can boost the performance of the Mask R-CNN on the test set of Unseen-VIS without losing performance on COCO classes.

To summarize our contributions:
\begin{itemize}[leftmargin=*]
\item We propose a Bayesian method to generate high-quality masks on unlabeled videos containing unseen classes;
\item We create a benchmark to evaluate the quality of generated masks on unlabeled videos;
\item We demonstrate on our benchmark that our proposed method can be used to improve the performance of an instance segmentation model on unseen classes.
\end{itemize}


\section{Related Work}
\label{sec:related_work}

In this section, we first review recent works on instance segmentation from color images, especially those targeting new classes. We also review self-learning methods on unlabeled images. Finally, we review several works for video object segmentation as this topic is closely related to our approach.

\paragraph{Image-level Instance Segmentation.}

While state-of-the-art methods for object detection and segmentation~\cite{he2017mask, tian2020conditional, wang2020solo} rely on large amounts of manually labeled images, a few works aim at reducing the annotation burden for learning to detect and segment object classes.  They however still require manual annotations: Weakly-supervised methods require no mask annotations but bounding boxes annotations~\cite{hu2018learning, Kuo2019ShapeMaskLT, Zhou2020LearningSP} and few-shot methods require a (small) number of manual object masks~\cite{Fan2020FGNFG,Yan2019MetaRT}.

\paragraph{Self-Learning on Unlabeled Data.}

Recently, several methods have been proposed to explore self-supervision using unlabeled static images. They use data distillation~\cite{Radosavovic2018DataDT}, unlabeled images from the web~\cite{li2020improving}, consistency across image flipping~\cite{jeong2019consistency}, or an estimate of the uncertainty of prediction~\cite{neverova2019correlated}. While such approaches are very interesting, unlabelled videos are easily available and have the potential to make the results much more reliable. We compare our method against the most representative methods in our experiments and show we achieve much better performance.

Another type of approach proceeds in a bottom-up fashion by grouping pixels with similar colors or image features to generate masks~\cite{burgess2019monet,greff2019multi,slot_attention,pham2018bayesian,wong2020identifying}. However, this can be easily affected by local textures or colors, and some of these methods have been demonstrated on synthetic images only. Because it starts from a pre-trained instance segmentation model, our approach is much more robust.

Like us, some methods leverage unlabeled videos for on urban scene segmentation~\cite{Chen2020LeveragingSL} and face and human detection\cite{jin2018unsupervised}. However, these works only focus on how to enhance the model performance on the existing classes and do not consider novel classes.  \cite{ovsep2019large} makes use of stereo video data together with depth information to reconstruct a static background, then object proposals are generated from the foreground regions by subtraction.  While this is an interesting approach, they require depth data and static backgrounds.

\paragraph{Video Object Segmentation.}

Our  work  also  relates  to   One-Shot,  Zero-Shot  video  object
segmentation~(VOS) and saliency-based video object segmentation. One-Shot VOS aims at segmenting the objects in the video when the ground truth segmentation is given for a frame. One-Shot VOS methods typically warp the provided segmentation to other frames~\cite{caelles2017one, uvc_2019, wang2019learning}. They thus require manual  annotations and cannot generate new predictions if new objects appear. Some Zero-Shot methods~\cite{cheng2017segflow, oh2019video, ventura2019rvos, voigtlaender2019feelvos} are trained with video labels on seen classes and are able to generalize to unseen classes, but video labeling is very labor-extensive. Some methods look for salient regions in videos~\cite{Chen_TransferSeg_2018, croitoru2017unsupervised, dave2019towards, faktor2014video, jain2017fusionseg, Koh2017Primary, lu2020learning, papazoglou2013fast}, as salient regions tend to correspond to objects. However, saliency prediction has two major limitations for our purpose: (a) It can be fooled by non-salient camouflaged objects. (b) Two adjacent objects would be merged into a single salient region, while we want to identify them individually.

Like us, a few methods already adopt a proposal-based approach~\cite{bergmann2019tracking, luiten2020unovost, milan2015joint}, but rely on classical tracking algorithms to track the proposals such as tracklets. By contrast, our approach relies on (non-realistic) analysis-by-synthesis. Analysis-by-synthesis is an old concept in computer vision, but has been recently increasingly popular. By aiming at explaining the whole image, it can exploit more information. Moreover, it is conceptually simple and requires few easy-to-fix hyper-parameters. We show in our experiments that our method performs better than the state-of-the-art video object segmentation method UnOVOST~\cite{luiten2020unovost} for the purpose of generating object masks.


\section{Method}
\label{sec:method}
As discussed in Section~\ref{sec:introduction}, our goal is to improve the performance of a pre-trained class-agnostic instance segmentation on unseen classes. Our pipeline consists of three steps:
\begin{itemize}[leftmargin=*]
    \item Mask Generation: We use our baseline instance segmentation network on unlabeled videos containing unseen classes for mask generation;
    \item Mask Selection: We apply our method to automatically select the correct masks on unlabeled videos;
    \item Model Refinement: We use our generated masks to fine-tune or retrain our baseline network to boost its performance on unseen classes.
\end{itemize}

In this section, we present our baseline instance segmentation network and our approach for automatically selecting high-quality masks by exploring the video information. As we will show in the following section, compared to  exhaustive search, our approach is highly efficient and requires few easy-to-fix hyper-parameters.

\subsection{Baseline Network for Mask Generation}
\label{sec:baseline}
To generate masks from unlabeled videos, we use a class-agnostic Mask R-CNN~\cite{he2017mask} with a ResNet-50-FPN~\cite{lin2017feature} backbone as our baseline network. Following previous work~\cite{ovsep20204d}, we refer to this class-agnostic Mask R-CNN as 'MP~R-CNN' for Mask Proposal R-CNN, as it aims only at generating mask proposals regardless of object classes. Note that in practice, Mask R-CNN could be replaced by any other trainable instance segmentation methods. As we mentioned in Section~\ref{sec:introduction}, the instance segmentation network may assign low confidence scores for some correct detections of unseen classes. Therefore, during the mask generation stage, we set the confidence score threshold to 0 to get as many detections as possible.

\subsection{Mask Selection}

Given a video of $T$ frames, we start from a set of mask candidates $\mathcal{\mask}_t = \{ \mask_{t,1}..\mask_{t,N}\}$ for each frame $\img_t$ obtained using our baseline network, with $N$ the number of masks candidates in $\img_t$. To select the mask candidates that actually correspond to objects, we exploit the following cues and constraint:
\begin{itemize}[wide=0pt]
\item The \textbf{``Background cue''}: Segmenting typical backgrounds such as sky or grass gives us a cue about where the objects are, whether they move or not.
\item The \textbf{``Flow cue''}: The optical flow between consecutive frames gives us a cue about the moving objects. 
\item The \textbf{``Consistency cue''}: The selected masks should be consistent not only between consecutive frames, but also over long sequences.
\item The \textbf{``Non-overlapping constraint''}: An additional constraint that is usually overlooked is that the masks should not overlap: Ideally, one pixel in the image can belong to at most one mask. 
\end{itemize}

As we will show in the following sections, each cue corresponds to a loss term in the final objective function. Each of them and the non-overlapping constraint contribute to removing the false positives, as will be demonstrated by our ablation study in Section~\ref{sec:ablation}.

To combine these cues to select the correct masks in a given video sequence, we rely on a Bayesian framework. This selection problem can be formalized as maximizing the probability of the detected masks given the frames of the video:
\begin{equation}
     P\big( \comb_1 ,.., \comb_T \given \img_1 ,.., \img_T \big) \> , 
\label{eq:pb1}
\end{equation}
where $\comb_t$ is a set of binary random variables, with $\comb_{t,i} = 1$ corresponding to the event that mask $\mask_{t,i}$ is selected and 0 that it is not. We show in the supplementary material that maximizing this probability is equivalent to minimizing the following objective function:
\begin{equation}
\begin{array}{l}
\argmin_{ \{\Delta_1,..,\Delta_T\} } \quad \sum_t \big( \lambda_I \calL_I\big(\img_t, \Delta_t) \; +  \\[1mm]
\quad \quad \quad \quad \quad \quad \quad \lambda_F \calL_F\big(\flow_t, \img_t, \img_{t+1}, \Delta_t, \Delta_{t+1} \big) \; + \\[1mm]
\quad \quad \quad \quad \quad \quad \quad \lambda_p \calL_p (\Delta_t, \Delta_{t+1}) \big) \> ,
\label{eq:argmin_final}
\end{array}
\end{equation}
under the non-overlapping constraint that will be detailed below. $\lambda_I$, $\lambda_F$, and $\lambda_p$ are constant weights. $\flow_t$ represents the optical flow for the pair of frames $(I_t, I_{t+1})$. $\Delta_t = \{ \delta_{t,\!1} ,.., \delta_{t\!,N}\!\}$ denotes the realization of $\comb_t$ where $\delta_{t,i}$ is the realization of the random binary variable $\comb_{t,i}$. $\delta_{t,i} = 1$ when $\mask_{t,i}$ is selected, otherwise $\delta_{t,i} = 0$. 


We call $\calL_I$ the Background loss and $\calL_F$ the Flow loss, as they exploit the Background cue and the Flow cue respectively. $\calL_p$ enforces consistent selections between consecutive frames. We detail these three losses below.


\subsubsection{Background Loss $\calL_I$}

\begin{figure}
\begin{center}
  \begin{tabular}{ccc}
    \includegraphics[width=1.0\linewidth]{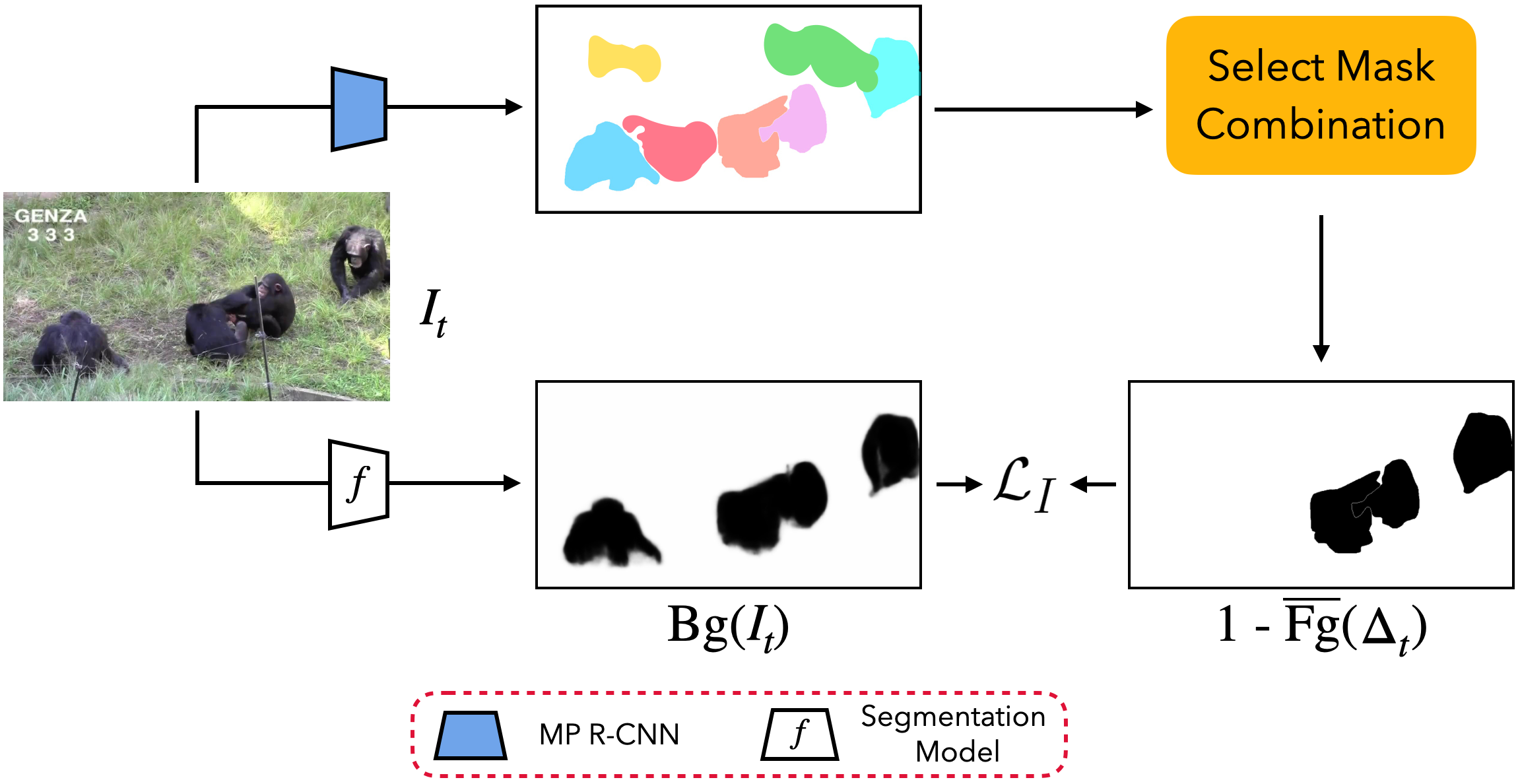}
\end{tabular}
\end{center}
\caption{\label{fig:LI} To evaluate the background loss $\calL_I$ of Eq.~\eqref{eq:LI}, we compare the background predicted for the image and the background of the selected masks.}
\vspace{-3mm}
\end{figure}

We use $\calL_I$ to exploit the Background cue that hints at where the objects are.  As shown in Figure~\ref{fig:LI}, to evaluate it, we compare a binary image generated for the selected masks and the foreground/background probability map predicted by a binary segmentation network $f$ by calculating their cross entropy, as the image background should match the background of the selected masks.  By doing this comparison over all the image locations, we can exploit information from the whole image to guide the mask selection---we will rely on the same strategy for the other terms. Formally, we take
\begin{equation}
 \calL_I\big(\img_t, \Delta_t\big) = \text{CE}\big(\Bg(I_t), 1 - \syntFg(\Delta_t)\big) \> ,
 \label{eq:LI}
\end{equation}
\noindent where $\text{CE}$ denotes the cross-entropy, $\Bg(I_t)$ is a probability map for each pixel to belong to the background as predicted by the network $f$, and $\syntFg(\Delta_t)$ is the binary image of masks in $\mathcal{\mask}_t$ such that $\delta_{t,i} = 1$. For $f$, we use the network architecture proposed in \cite{kirillov2019panoptic} trained on the same training data as our baseline network for mask generation. The details about the segmentation network $f$ can be found in the supplementary material.

\subsubsection{Flow Loss $\calL_F$}

We use $\calL_F$ to exploit the Flow cue: The optical flow is the result of the object motions and the camera motion, and thus also hints at where the objects are.  Even if an object is static but the camera is in motion, when the distance between the camera and the background is large enough, relative motion will make the optical flow of the object regions stand out from the background optical flow.

\begin{figure}
\begin{center}
  \begin{tabular}{cccc}
    \includegraphics[width=1.0\linewidth]{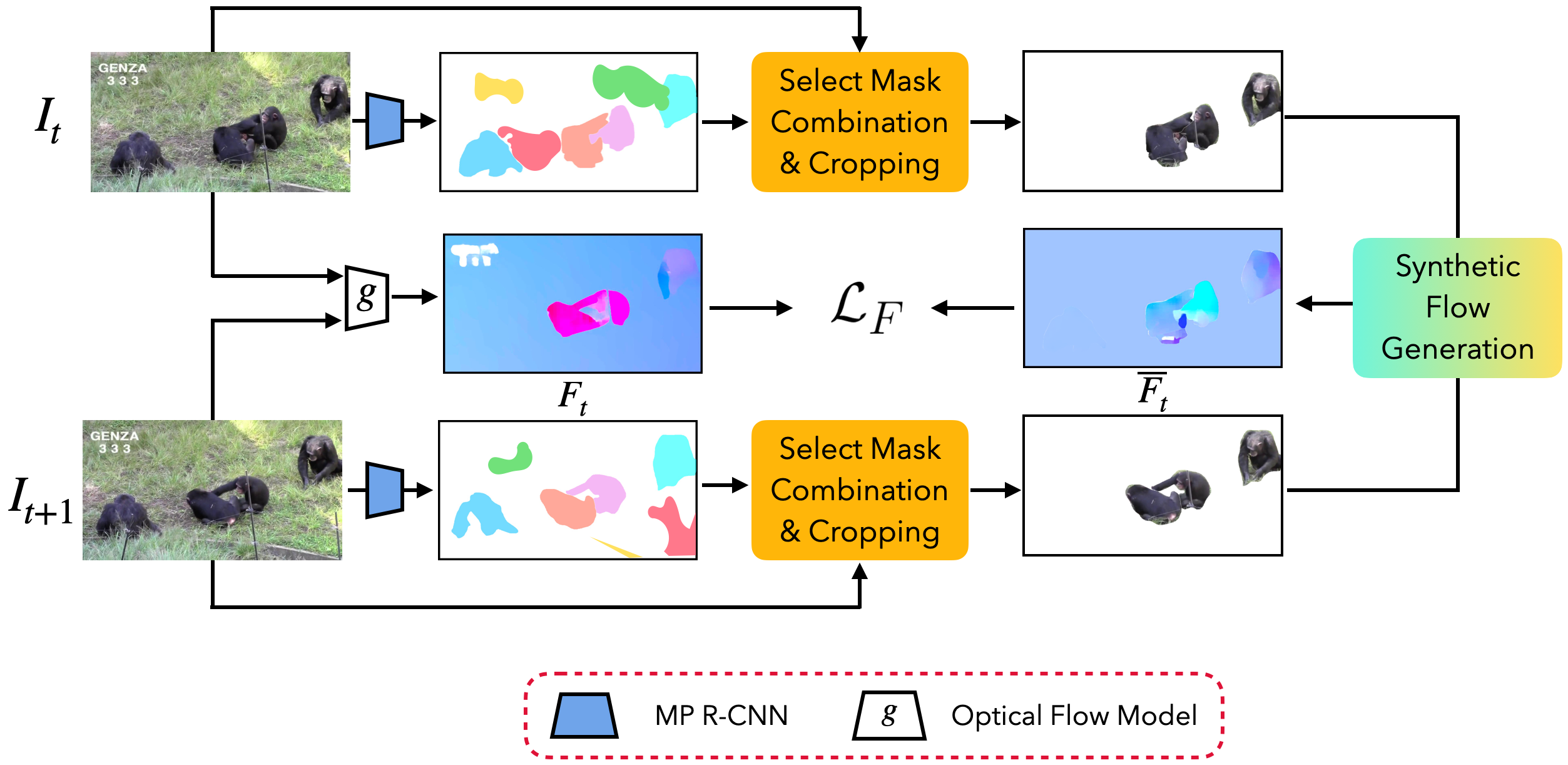}
\end{tabular}
\end{center}
\caption{\label{fig:LF} To evaluate the flow loss $\calL_F$ of Eq.~\eqref{eq:LF}, we compare the optical flow estimated between two consecutive images and the optical flow computed for the masks selected in the two images.}
\end{figure}

Figure~\ref{fig:LF} shows how we evaluate this term.  We compare the flow predicted by an optical flow estimator $g$ and a ``synthetic optical flow'' generated using the masks selected in $\mathcal{\mask}_t$ and $\mathcal{\mask}_{t+1}$. Similar to the $\calL_I$ term, this comparison allows us to exploit information from all the image locations. In practice, we use the method of \cite{teed2020raft} for $g$. To generate the synthetic optical flow, we use the colors of the pixels in the selected masks to compute their optical flow.  We average the flow in $\flow_t$ on the pixels that do not belong to any mask to assign these pixels the resulting value. The detailed procedure can be found in the supplementary material.
 
Using this procedure, the measured flow $\flow_t=g(I_t,I_{t+1})$ and the synthetic flow are similar when all moving objects are correctly selected in both frames, even when the camera is in motion.  More formally, we take:
\begin{align}
  & \> \calL_F\big(\flow_t, \img_t, \img_{t+1}, \Delta_t, \Delta_{t+1} \big) = \big\|\flow_t - \syntflow_t \big\|_1 \> ,
\label{eq:LF}
\end{align}
where $\syntflow_t = \syntflow_t(\img_t, \img_{t+1}, \Delta_t,\Delta_{t+1})$ is the synthetic flow generated for the selected masks in $\mathcal{\mask}_t$ and $\mathcal{\mask}_{t+1}$.  We use the L1-norm to compare the two flows to be robust to outlier values that are very common in the predicted flow.  Figure~\ref{fig:LF} shows examples for $\flow_t$ and $\syntflow_t$.

\subsubsection{Regularization Loss $\calL_p$ and Constraint}
\label{sec:constraints} 

\begin{figure}
\begin{center}
  \begin{tabular}{cc}
   \includegraphics[width=1.0\linewidth]{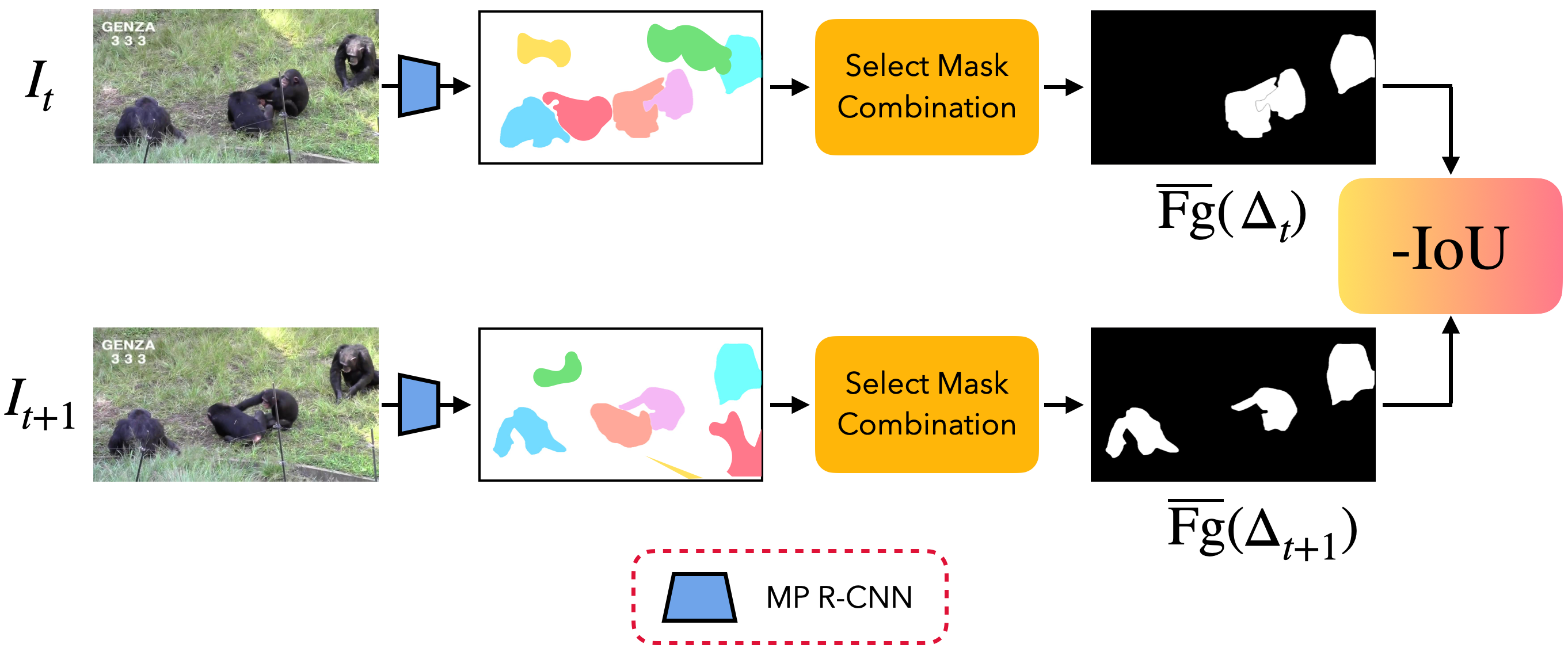}
\end{tabular}
\end{center}
\caption{\label{fig:LP} To evaluate the regularization loss $\calL_p$ of Eq.~\eqref{eq:Motion}, we compare the binary images of the selected masks in two consecutive images.}
\vspace{-3mm}
\end{figure}


As discussed above, the optimization in Eq.~\eqref{eq:argmin_final} should be done under the constraint that no masks selected for the same frame overlap each other.

$\calL_p\big(\Delta_t, \Delta_{t+1}\big)$ is usually interpreted in tracking problems as a motion model. We use it to enforce a consistent selection of masks between consecutive frames. Figure~\ref{fig:LP} shows how we compute it: We use a very simple motion model and assume that the objects move slowly, in other words, the areas segmented as objects do not change abruptly between two consecutive frames. Formally, we take :
\begin{equation}
\label{eq:Motion}
 \calL_p\big(\Delta_t, \Delta_{t+1}\big) = -\mathrm{IoU}\big(\syntFg(\Delta_t), \syntFg(\Delta_{t+1})\big) \> ,
\end{equation}
\ie the negative Intersection-over-Union between the binary images of the masks selected for frames $\img_t$ and $\img_{t+1}$. It is set to 0 when no masks are selected for none of the two images.


\subsection{Two-Stage Optimization}
\label{sec:two-stage-opt}

In this section, we introduce an efficient way to minimize  Eq.~\eqref{eq:argmin_final}. Note that minimizing this function requires to optimize on all the frames simultaneously. A naive approach is to apply exhaustive search for the solution of the problem, where the number of evaluations of the objective function would be $O(2^{NT)}$, with $N$ the number of mask candidates per frame and $T$ the number of frames (typical values $N=15$ and $T=180$ would require $\sim 10^{810}$ evaluations). This is clearly computationally prohibitive.

\begin{figure}
\begin{center}
  \begin{tabular}{cc}
   \includegraphics[width=1.\linewidth]{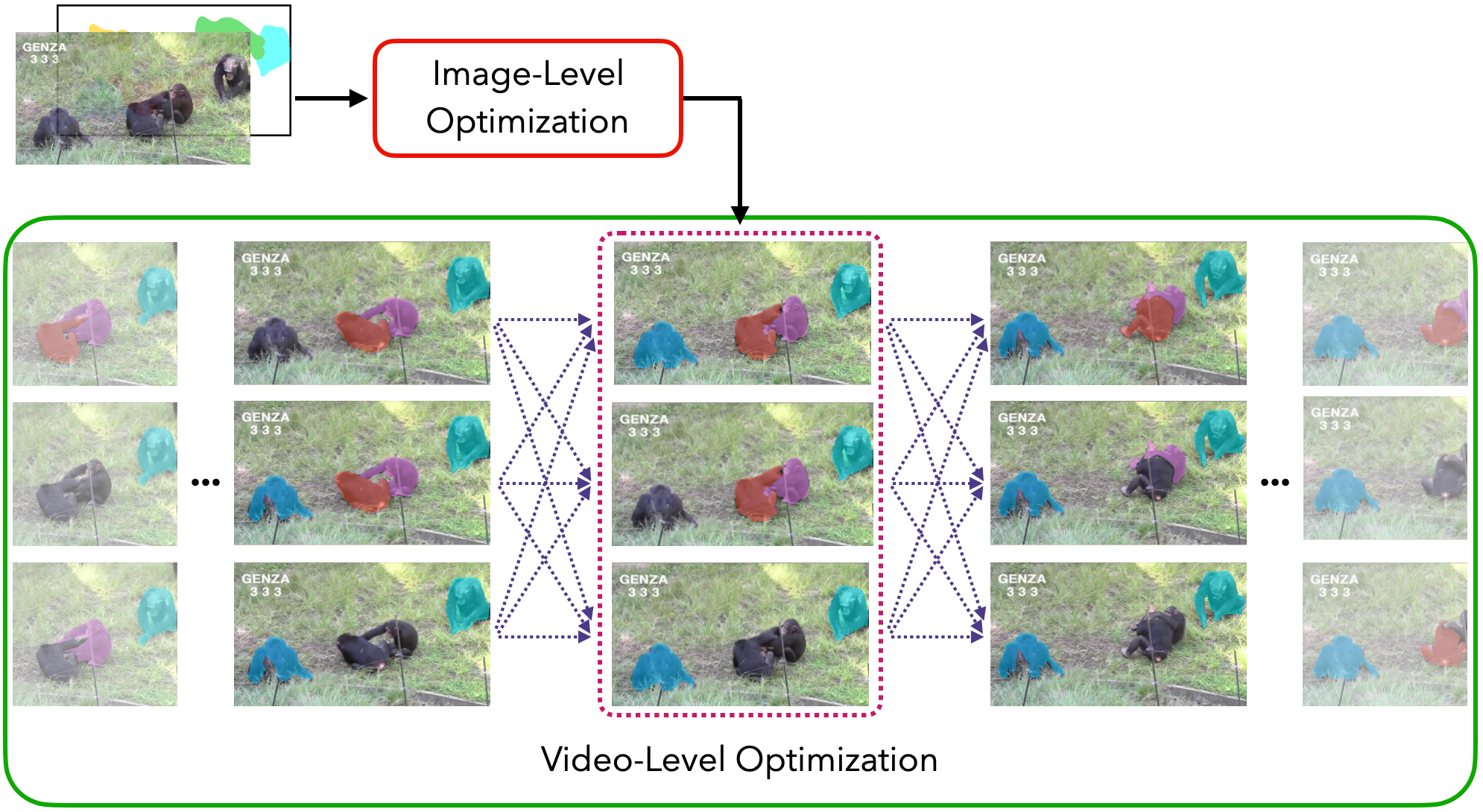}
\end{tabular}
\end{center}
\caption{\label{fig:VG} Given a video, we first run the image-level optimisation on each frame and get the top-$K$ combinations of masks for each frame. The video-level optimisation selects the best combination for each frame efficiently by solving a shortest path problem. For this figure, $K$ is set to 3. More details are given in Section~\ref{sec:two-stage-opt}.}
\vspace{-3mm}
\end{figure}

We provide here an efficient two-stage algorithm, depicted in Figure~\ref{fig:VG}. In the first stage, based on the background loss $\calL_I$ and the non-overlapping constraint, we select the top-$K$ most promising combinations of masks for each frame independently. Note that the combinations violating the non-overlapping constraint are simply discarded. Then, in the second stage, we optimize the complete objective function over all frames simultaneously to find the best combinations for each frame. For both stages, we can use Dijkstra's algorithm~\cite{dijkstra1959note} to significantly decrease the complexity of the computations. In the worst case, the number of evaluations of the objective function becomes $O(KTN^3+K^2T^2)$. We use $K=10$ in practice, which reduces the required evaluations from $\sim 10^{810}$ to $\sim 10^{7}$ for the numerical example above.

This optimization problem is related to many previous works on multiple object tracking~\cite{berclaz2011multiple, ma2012maximum, Perazzi_2015_ICCV, wu2011efficient, zamir2012gmcp, zhang2013video}, which typically use graph-based methods to solve related problems efficiently. One of the main differences with these works is that, in our video-level optimization, each node corresponds to a combination of masks instead of a single bounding box or mask. Besides, we do not have access to the object classes, and our optimization is under the constraint that the masks do not overlap, while these works typically rely on bounding boxes that can overlap when the objects are close to each other.

\paragraph{1. Image-Level Optimization. }

At this stage, for each frame $I_t$, we look for the top-$K$ combinations of masks in the power set $\mathcal{P}(\mathcal{\mask}_t)$ that minimize the Eq.\eqref{eq:LI} under the non-overlapping constraint. An exhaustive search would take $2^N$ evaluations of the objective function.

However, we note that this problem can be formulated as a K-shortest path search problem in a binary tree, where each pair of branches of a node corresponds to the selection or not of a mask, and each branch has an associated weight: This weight is set to infinity if the branch corresponds to the selection of a mask that overlaps with one of its ancestors, otherwise it depends on the value of $\calL_I$ computed only on the mask.  By iteratively applying Dijkstra's algorithm to find the top-$K$ combinations of masks, we reduce the number of evaluations to $O(KN^3)$. Note that our proposed algorithm is agnostic to the order of the mask candidates in $\mathcal{\mask}_t$. More details can be found in the supplementary material.

\paragraph{2. Video-Level Optimization. } 

As shown  in Figure~\ref{fig:VG}, we generate a graph with the remaining top-$K$ combinations of masks for each frame: Each node corresponds to one combination and each edge is labeled with the loss given in Eq.\eqref{eq:argmin_final} for the two combinations it links. Finding the best combination for each frame becomes the problem of finding the shortest path in this graph. Instead of doing $2^{K T}$ evaluations of the objective function to find the shortest path, we can use Dijkstra's algorithm~\cite{dijkstra1959note} here as well to significantly accelerate the speed of our algorithm. The number of evaluations is reduced to $O(K^2T^2)$. We explain in more detail how we build the graph in the supplementary material.

\section{Experiments}
\label{sec:exp}
In this section, we first present our benchmark for evaluating our approach. Then we compare our method with several previous methods for the purpose of generating masks on unlabeled videos and analyze the results.  We also conduct a thorough ablation study to show the influence of different components of our method.  Due to space limits, implementation details were moved to the supplementary material.

\subsection{Experimental Setup}
\label{sec:setup}

\noindent\textbf{Implementation Details.} 
As described in Section~\ref{sec:baseline}, we use a class-agnostic Mask R-CNN~\cite{he2017mask} with a ResNet-50-FPN~\cite{lin2017feature} backbone as our baseline for mask generation. Our baseline is pre-trained on the COCO dataset, which contains 80 classes and 115k training images.  We follow the training strategy as described in \cite{he2017mask}. We take $\lambda_I = \lambda_F = 1$ and $\lambda_p = 0.5$ for the weights in Eq.~\eqref{eq:argmin_final} in all our experiments.

\noindent
\textbf{Evaluation.} To benchmark our method, we created a dataset we call ``Unseen-VIS''. The training part of ``Unseen-VIS'' consists of videos collected from the YouTube Video Instance Segmentation~(YouTube~VIS)~\cite{yang2019video} and is used for mask generation. The test part of ``Unseen-VIS'' contains static images extracted from YouTube~VIS for evaluation.

The original YouTube~VIS dataset contains 2,883 videos with 131k object instances spanning 40 classes, among which 24 coincide with COCO. We thus consider the remaining 16 classes~\footnote{panda, lizard, seal, shark, mouse, frog, tiger, leopard, fox, deer, ape, snake, monkey, rabbit, fish, turtle.} as {\it unseen} classes which results in 795 videos in total. We randomly selected 595 videos as the training set, which we refer as Unseen-VIS-train. The labeled static images in the remaining 200 videos are used for evaluation, which we refer as Unseen-VIS-test. All the videos of Unseen-VIS-train are used as unlabeled videos, and their ground truth masks are ignored.

For quantitative evaluation, we rely on the standard COCO metrics: $AP$, $AP_{50}$, $AP_{75}$, and $ AR_{1} $, $ AR_{3} $ and $ AR_{5} $ as the maximum number of objects per image in our testing set is 4. We do not use $ AP_{S} $, $ AP_{M} $, and $ AP_{L}$ as the object scales in COCO differ largely from YouTube~VIS.

\newcommand{\PreserveBackslash}[1]{\let\temp=\\#1\let\\=\temp}
\newcolumntype{C}[1]{>{\PreserveBackslash\centering}p{#1}}
\newcolumntype{R}[1]{>{\PreserveBackslash\raggedleft}p{#1}}
\newcolumntype{L}[1]{>{\PreserveBackslash\raggedright}p{#1}}
\newcommand{\myparagraph}[1]{\noindent {\bf #1}}

\subsection{Results on Unseen-VIS-test}
\label{sec:unseen_seg}

\subsubsection{Video-Annotation-Free Mask Generation}

We first compare our method to other approaches that can also generate masks given video sequences without using any video annotations. Each method is first applied on the Unseen-VIS-train dataset for mask generation, then we compare the performance of MP~R-CNN on the Unseen-VIS-test set after fine-tuning on these masks.  As only one over five frames is annotated in Unseen-VIS-train~(19352 annotated frames in total), we thus use only the masks of these frames for training for fair comparison among different methods.

\myparagraph{Saliency/Flow-based methods.} FST~\cite{papazoglou2013fast} and NLC~\cite{faktor2014video} can generate masks from videos by estimating the saliency and motion of the objects in videos. IOA~\cite{croitoru2017unsupervised} trains a deep neural network on the output of an unsupervised soft foreground segmentation algorithm~\cite{stretcu2015multiple} to segment objects in videos. These methods can identify  moving regions in the videos but can not separate adjacent objects in the images, thus a proposal generated by these methods may actually correspond to several objects.

\myparagraph{Tracking-based methods.} TWB~\cite{bergmann2019tracking} and UnOVOST~\cite{luiten2020unovost} rely on a frame-by-frame tracking pipeline applied to mask proposals. These methods are the closest methods to ours as we all rely on an instance segmentation model for proposal generation. However, as we mentioned before, for our final goal (training a better object detector), we do not need to keep track of the detected objects.

\myparagraph{Similarity Propagation.} Given masks for a frame, UVC~\cite{uvc_2019} warps these masks to consecutive frames based on the estimated correspondences between consecutive frames. For this experiment, we use it in a zero-shot setting ("ZS-UVC"), where the masks of the first frame are instead generated by thresholding the confidence score on the first frame prediction of MP~R-CNN~(we use a threshold of $0.1$ in practice).

\myparagraph{Self-training methods.} We also compare our method with the self-learning Data Distillation~(DD) method~\cite{Radosavovic2018DataDT}. We follow their proposed test time augmentation to generate masks on each frame of the Unseen-VIS-train videos independently, as this method performs on single images.

\begin{table}[!t]
  \addtolength{\tabcolsep}{1pt}
  \begin{center}
    \scalebox{.85}{
    \begin{tabular}{@{}l | ccc ccc @{}}
	\toprule
	Method used for & \multicolumn{6}{c}{Unseen-VIS-test}\\
	mask generation & $AP$ & $AP_{50}$ & $AP_{75}$ & $AR_1$ & $AR_3$ & $AR_{5}$ \\ 
	\midrule
	(bef. fine-tuning) & 35.8 & 61.2 & 38.1 & 33.3 & 47.3 & 50.3 \\
	\hline
	NLC~\cite{faktor2014video} &  1.2 &  3.8 &  1.0 &  2.4 & 5.3 & 6.9 \\
	IOA~\cite{croitoru2017unsupervised} & 2.4 & 8.5 & 0.9 & 6.9 & 8.7 & 9.5 \\
	FST~\cite{papazoglou2013fast} & 17.0 & 41.8 & 11.3 & 22.0 & 30.6 & 33.1 \\
	UnOVOST~\cite{luiten2020unovost} & 31.1 & 55.6 & 32.2 & 29.9 & 44.5 & 48.2 \\
	TWB~\cite{bergmann2019tracking} & 31.2 & 53.4 & 32.8 & 31.5 & 46.7 & 50.0 \\
	DD~\cite{Radosavovic2018DataDT} & 36.6 & 63.8 & 38.5 & 32.5 & 46.2 & 49.3 \\
	ZS-UVC~\cite{uvc_2019} & 21.2 & 42.6 & 19.9 & 26.3 & 40.0 & 43.2 \\
	\textbf{Ours} &  \textbf{39.0} & \textbf{67.9} & \textbf{41.3} & \textbf{35.2} & \textbf{48.9} & \textbf{51.4} \\
	\bottomrule
    \end{tabular}
    }
  \end{center}
  \caption{
    {\bf Mask Generation without Video Annotation.}
    Performance of MP R-CNN on Unseen-VIS-test after fine-tuning on masks generated by various methods applied to Unseen-VIS-train without using any video annotation. 
  }
  \label{tab:comparison table wo video annos}
  \vspace{-3mm}
\end{table}

We report the results in Table~\ref{tab:comparison table wo video annos}. Compared with the Saliency/Flow-based methods and Tracking-based methods, the masks generated by our method are of high quality and can largely improve the performance of the baseline network in all metrics. Our method also outperforms the state-of-the-art self-training method~(DD)~\cite{Radosavovic2018DataDT} for two-stage object detection by a large margin.

\subsubsection{Video-Annotation-Dependent Mask Generation}

In addition to the aforementioned methods, we further compare with two methods that adopt different settings and explore the upper bound of our method.

\myparagraph{RVOS. } RVOS~\cite{ventura2019rvos} is an end-to-end video object segmentation framework that directly runs on videos, which requires labeled videos for training. We adopt the Zero-Shot setting for RVOS, where it is trained with ResNet50~\cite{he2016deep} as backbone on 1089 videos~(25869 annotated frames in total) of seen classes of YouTube~VIS and directly applied to the Unseen-VIS-train videos for mask generation.

\myparagraph{OS-UVC. } Here, we consider the One-Shot setting for UVC, where the ground truth masks of the first frame are given for all Unseen-VIS-train videos.

\begin{table}[!t]
  \addtolength{\tabcolsep}{-2.8pt}
  \begin{center}
    \scalebox{.75}{
    \begin{tabular}{@{} l | cc | ccc ccc @{}}
	\toprule
	Method used for & \multicolumn{2}{c|}{Video Annotations} & \multicolumn{6}{c}{Unseen-VIS-test}\\
	mask generation & Seen & Unseen & $AP$ & $AP_{50}$ & $AP_{75}$ & $AR_1$ & $AR_3$ & $AR_{5}$ \\ 
	\midrule
	(bef. fine-tuning) & & & 35.8 & 61.2 & 38.1 & 33.3 & 47.3 & 50.3 \\
	\hline
	RVOS~\cite{ventura2019rvos} & \checkmark & & 38.5 & 68.9 & 38.0 & 35.4 & 49.5 & 52.8 \\
	Ours & - & - & 39.0 & 67.9 & 41.3 & 35.2 & 48.9 & 51.4 \\
	OS-UVC~\cite{uvc_2019} & & \checkmark & 41.5 & 73.7 & 42.9 & 39.1 & 52.7 & 54.9 \\
	Selected using GT & & \checkmark & 42.7 & 75.1 & 45.3 & 37.3 & 53.4 & 53.6 \\
	Trained with GT & & \checkmark & \textcolor{mygray}{50.8} & \textcolor{mygray}{80.9} & \textcolor{mygray}{54.6} & \textcolor{mygray}{43.6} & \textcolor{mygray}{58.6} & \textcolor{mygray}{60.6} \\
	\bottomrule
    \end{tabular}
    }
  \end{center}

  \caption{ {\bf Mask Generation with Video Annotations.}  Performance of MP R-CNN on Unseen-VIS-test after fine-tuning on the masks generated by various methods that require manual video annotations except ours.  RVOS uses labeled videos of Seen classes for training. OS-UVC uses the ground truth masks for the first frame for mask generation.  "Selected using GT" represents the masks generated by MP R-CNN selected using ground truth masks and can be regarded as an upper-bound.  }
  \vspace{-3mm}
  \label{tab:comparison table w video annos}
\end{table}

\myparagraph{Selected using GT.} We use the ground truth mask labels of Unseen-VIS-train to select the masks predicted by MP R-CNN. The similarity among masks is evaluated based on their Intersection-over-Union,  and the Hungarian algorithm is used to select the masks that best match with the ground truth. This can be regarded as an upper bound that we can achieve given the predictions of MP~R-CNN.

\myparagraph{Trained with GT.} We report the performance of MP~R-CNN fine-tuned with ground truth mask labels of Unseen-VIS-train. This can also be regarded as an upper bound, where all the classes have already been Seen.

\begin{figure*}[!t]
    \centering
    \includegraphics[width=1.\textwidth]{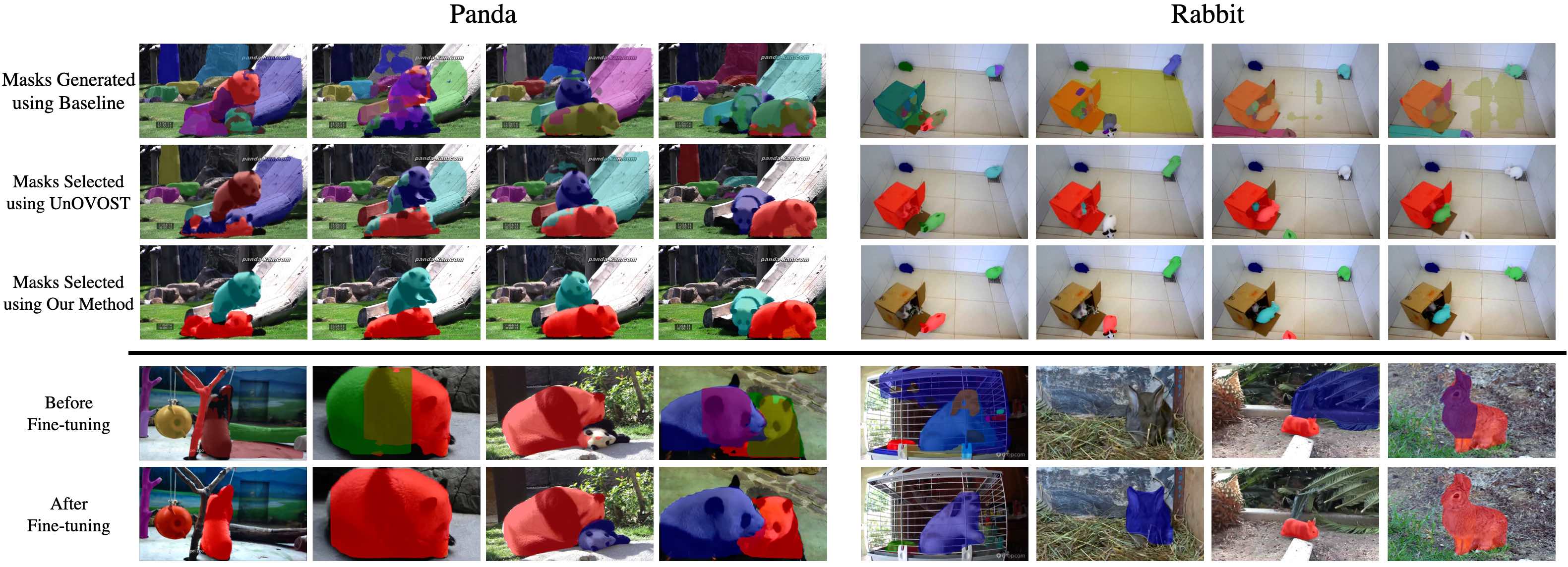}
    \vspace{-3mm}
    \caption{
      {\bf Qualitative results of selected masks on Unseen-VIS-train and detections of new classes on Unseen-VIS-test after fine-tuning on these selected masks.} 
{\bf Top:} First row: Masks detected by our baseline network MP~R-CNN on two sequences from Unseen-VIS-train; Second row: Masks selected by UnOVOST~\cite{luiten2020unovost}; Third row: Masks selected by our approach. Note that we keep the masks for the pandas and rabbits, and reject the masks that do not correspond to real objects. {\bf Bottom:}  Masks detected in still images from Unseen-VIS-test. Fourth row: Masks detected by MP~R-CNN before we fine-tuned it on the masks selected by our approach on Unseen-VIS-train; Fifth row: Masks detected by MP~R-CNN \emph{after} fine-tuning.  The masks generated by our method results in a significantly better model for the new classes: We can now correctly segment pandas and rabbits in new videos, even if no manual segmentations for pandas and rabbits were provided. \emph{For more examples, please refer to the video in the supplementary material. } 
    }
    \label{fig:my_label}
    \vspace{-4mm}
\end{figure*}

We report the results in Table~\ref{tab:comparison table w video annos}. Compared to  RVOS~\cite{ventura2019rvos} trained with labeled videos, our method can still achieve comparable results on recall while surpassing their results by a large margin on $AP_{75}$, 
which means that the masks selected by our method are of better quality.
Importantly for practical applications, our method was able to deal with the large domain gap between the images in COCO on which we pre-train MP~R-CNN and the frames in YouTube~VIS on which we apply and evaluate our method. RVOS was trained and applied on videos from YouTube~VIS, and therefore was not confronted to a domain gap.

While OS-UVC~\cite{uvc_2019} achieves higher results than our approach, it relies mainly on a high-quality first frame mask: We observe a large performance drop when we replace the ground truth masks ("OS-UVC" in Table~\ref{tab:comparison table w video annos}) by the predicted masks ("ZS-UVC" in Table~\ref{tab:comparison table wo video annos}).
Besides, it can only track objects visible in the first frame as it does not handle the emergence of new objects.

After simply retraining from scratch both on the 80 seen classes of COCO and the masks generated by our approach on Unseen-VIS-train, MP~R-CNN achieves 35.2 mask $AP$ on COCO minival and 38.9 mask $AP$ on Unseen-VIS-test. Compared with the MP~R-CNN pre-trained only with COCO dataset, which achieves 35.3 mask $AP$ on COCO minival, we achieve better performance on Unseen-VIS-test while maintaining the performance on COCO. 
More details can be found in the supplementary material.

\begin{table}[!t]
  \addtolength{\tabcolsep}{-2pt}
  \begin{center}
    \scalebox{.65}
	     {
	       \begin{tabular}{@{}l | c | c | ccc | ccc @{}}
	         \toprule
	         & {\it Use Unla-} & \multicolumn{1}{c|}{$\mathcal{J}\And\mathcal{F}$} & \multicolumn{3}{c|}{$\mathcal{J}$} & \multicolumn{3}{c}{$\mathcal{F}$}\\
	         Method & {\it beled Data} & \textbf{Mean} & \textbf{Mean} & \textbf{Recall} & \textbf{Decay} & \textbf{Mean} & \textbf{Recall} & \textbf{Decay} \\ 
	         \midrule
	         UnOVOST~\cite{luiten2020unovost} & & 56.2 & 54.4 & 63.7 & -0.01 & 57.9 & 65.0 & 0.00\\
	         UnOVOST+ & \checkmark & \textbf{59.9} & \textbf{59.1} & \textbf{70.0} & \textbf{-0.06} & \textbf{60.8} & \textbf{70.8} & \textbf{-0.03} \\
	         \bottomrule
	       \end{tabular}
    }
  \end{center}
  \vspace{-3mm}
  \caption{{\bf Zero-Shot Video Object Segmentation evaluation on the DAVIS dataset~\cite{perazzi2016benchmark}.}  UnOVOST~\cite{luiten2020unovost} relies on an instance segmentation network for mask generation.  UnOVOST+ row: After fine-tuning its mask generation network on the DAVIS training dataset using the masks generated by our approach, it achieves higher results on all metrics.}
  \label{tab:davis comparison table}
  \vspace{-4mm}
\end{table}
\vspace{-3mm}
\paragraph{Application to Zero-Shot Video Object Segmentation.} As one of the state-of-the-art zero-shot video object segmentation methods~$\footnote{https://davischallenge.org/challenge2019/leaderboards.html}$, UnOVOST~\cite{luiten2020unovost} segments the objects in the videos by linking the masks predicted by an instance segmentation network on each frame. As show in Table~\ref{tab:davis comparison table}, by fine-tuning the original mask generation model on the masks generated on the DAVIS training dataset~\cite{perazzi2016benchmark} using our method, we achieve much better results. This demonstrates that downstream tasks can benefit from the performance boost brought by our method.

\subsection{Ablation Study}
\label{sec:ablation}

Table~\ref{tab:overall ablation} shows the positive impact made by each loss term and constraint in Eq.~\eqref{eq:argmin_final}. The masks obtained by applying only the background loss $\calL_I$ can already improve the performance of baseline on unseen classes.  Similarly, adding the constraint that the masks should not overlap, the flow loss $\calL_F$, or the regularization loss $\calL_p$ has a positive impact.  In particular, this shows that both the flow loss $\calL_F$ and the regularization loss $\calL_p$ help reranking the combinations of masks given by the background loss.

\begin{table}[!t]
  \addtolength{\tabcolsep}{-1.pt}
  \begin{center}
    \scalebox{.78}
	     {
	       \begin{tabular}{@{}cccc | ccc ccc @{}}
	         \toprule
	         \textit{$\mathcal{L}_I$} & Overl.Constr. & \textit{$\mathcal{L}_F$} & \textit{$\mathcal{L}_p$} & $AP$ & $AP_{50}$ & $AP_{75}$ & $AR_1$ & $AR_3$ & $AR_{5}$ \\ 
	         \midrule
	          & & &  
	         & 35.8 & 61.2 & 38.1 & 33.3 & 47.3 & 50.3 \\
	         
	         \checkmark & & &
	         & 36.6 & 65.2 & 37.8 & 33.3 & 47.7 & 50.8 \\
	         
	         \checkmark & \checkmark & &
	         & 38.1 & 64.6 & 40.2 & 34.2 & 48.5 & 51.2 \\
	         
	         \checkmark & \checkmark & \checkmark & 
	         & 38.7 & 67.0 & 40.5 & 34.7 & 48.7 & 51.3 \\
	         
	         \checkmark & \checkmark & \checkmark & \checkmark
	         & 39.0 & 67.9 & 41.3 & 35.2 & 48.9 & 51.4 \\
	         \bottomrule
	       \end{tabular}
	     }
  \end{center}
  \vspace{-3mm}
  \caption{{\bf Ablation study} on the different components of our method.
  }
  \vspace{-4mm}
  \label{tab:overall ablation}
\end{table}


\section{Conclusion}

In this paper, we attacked the problem of localizing and segmenting objects from unseen classes without any manual mask labels. We showed that, based on an instance segmentation model pretrained on some seen classes, our method provides high-quality masks for unseen classes after analysing unlabeled videos, without requiring difficult-to-tune hyper-parameters. Moreover, we provided an efficient implementation by breaking down the computationally prohibitive optimization into a two-stage optimization.

It should nevertheless be noted that, in the unsupervised case, the concept of objects is quite ill-defined.  The boundary between ``things'' and ``stuff''~\cite{caesar2018coco, forsyth1996finding} is sometimes fuzzy: For example, should we consider a stone on the background as an object?  What if a person pushes this stone?  The granularity of the problem is also not clear: Should we consider a person as one object, or each of their clothes as individual objects?  This ambiguity does not arise in the supervised case, as the annotators decide what is an object, but makes the evaluation of unsupervised object detection difficult.  It may be important to rethink the definition of object, either by its shape or its function, to make the evaluation of unsupervised object segmentation more meaningful.

\vspace{-5mm}
\paragraph{Acknowledgement}
We thank Ze Chen, Zerui Chen, Changqian Yu, Enze Xie, Tianze Xiao, Yinda Xu, Xi Shen, Mathis Petrovich, and Philippe Chiberre for valuable feedback. This project has received funding from the CHIST-ERA IPALM project.

{\small
\bibliographystyle{ieee_fullname}
\bibliography{egbib}
}


\end{document}